\documentclass[preprint,12pt]{elsarticle}
\usepackage{float}
\usepackage{etoolbox}
\usepackage[noend]{algpseudocode}
\usepackage{algorithm}
\usepackage{algorithmicx}
\usepackage{graphicx} 
\usepackage{subcaption}
\patchcmd{\abstract}{Abstract}{Abstract}{}{}

\usepackage{graphicx}
\usepackage{amssymb}
\usepackage{amsmath}

\journal{ArXiv}

\begin{document}

\begin{frontmatter}


\title{Band Target Entropy Minimization and Target Partial Least Squares for Spectral Recovery and Calibration }

\author{Casey Kneale}
\author{Steven D. Brown\corref{cor1}}
\cortext[cor1]{sdb@udel.edu}
\address{Department of Chemistry and Biochemistry, University of Delaware, 163 the Green, Newark, DE, 19716, USA\fnref{label3}}


\begin{abstract}
The resolution and calibration of pure spectra of minority components in measurements of chemical mixtures without prior knowledge of the mixture is a challenging problem. In this work, a combination of band target entropy minimization (BTEM) and target partial least squares (T-PLS) was used to obtain estimates for single pure component spectra and to calibrate those estimates in a true, one-at-a-time fashion. This approach allows for minor components to be targeted and their relative amounts estimated in the presence of other varying components in spectral data. The use of T-PLS estimation is an improvement to the BTEM method because it overcomes the need to identify all of the pure components prior to estimation. Estimated amounts from this combination were found to be similar to those obtained from a standard method, multivariate curve resolution-alternating least squares (MCR-ALS), on a simple, three component mixture dataset. Studies from two experimental datasets demonstrate where the combination of BTEM and T-PLS could model the pure component spectra and obtain concentration profiles of minor components but MCR-ALS could not.

\end{abstract}


\end{frontmatter}


\section{Introduction}
\label{S:1}
Resolving poorly represented low intensity pure component spectra and concentration profiles from chemical mixtures without a priori knowledge is an open field of research. Provided that varying amounts of the target compound are present in the data matrix, many methods, such as self modeling curve resolution \cite{Lawton},evolving factor analysis \cite{maeder,efa}, window factor analysis \cite{wfa}, heuristic evolving latent projections \cite{help}, iterative target transformation factor analysis \cite{itffa}, simple-to-use interactive self-modeling mixture analysis (SIMPLISMA) \cite{SIMPLISMA}, and the standard multivariate curve resolution-alternating least squares (MCR-ALS) \cite{MCRorig} have been used to resolve the spectral signatures of major contributors from mixture data. However, when the component of interest is present at relatively low levels, the resolution of that component by the these methods is problematic because its spectral response can be lost in the contributions of noise \cite{chemicalrank} or its variation can be small enough relative to those of the other contributions that a given method will fail to resolve it from the larger components \cite{itffa}.  

Band target entropy minimization (BTEM) takes a different approach to the problem of spectral signature recovery. Band target entropy minimization estimates a single, pure component spectrum from a linear combination of weighted loadings obtained by principal component analysis \cite{BTEM}. This approach allows orthogonal contributions of variance to be combined to form a spectral estimate. One potential advantage with the use of BTEM is that the technique does not require that all components in a mixture have adequate variation in the samples, as long as the single target for resolution is represented in the PCA loadings \cite{BTEM, BTEMex}. This advantage cannot be attributed to many other methods because most methods attempt to simultaneously discover and resolve the spectral signatures of every component in a mixture \cite{Lawton,itffa,SIMPLISMA,MCRorig}. 

Although BTEM has been used for the recovery of individual components in a mixture, in order to obtain estimates of relative amounts, the resulting pure component spectral estimates have been used as a regression coefficients in classical least squares (CLS) models \cite{BTEM,BTEMex}. This approach poses a problem because classical least squares regression requires that all pure component responses embedded in a mixture be included in the regression model. If this condition is not met, the concentration estimate matrix that results from CLS has insufficient rank and cannot be used to discriminate the additive effects of all mixture components from a given target compound. Therefore, previous work using BTEM spectral recovery required either the discovery of all pure component responses for the mixture in order to obtain their approximate amounts, or that a regression is performed on spectra which have one component or on regions where only the available pure component spectra contribute. In many cases, an analyst may only be interested in the relative amount of one component for quantitation due to the presence of an undersampled or poorly represented background. Similarly, to characterize an unknown, minor component of a mixture, the entire spectral range may need to be considered. Thus, the potential advantage of being able to recover single component spectra from BTEM has not been utilized in a way that allows for determining the relative amounts of single components. 

It is well known that calibration methods such as partial least squares or principal component regression \cite{plsgeladi} can overcome issues associated with rank deficiencies that arise when the responses for all components of a mixture are not fully known. However, in curve resolution experiments, prior knowledge of property values is not available. Thus, the usual regression methods cannot be used because only the target spectrum and the spectral data are available. 

To avoid the problems associated with conventional regression methods, target partial least squares (T-PLS) \cite{TPLS} can be used for creating regression models built from single, pure component spectral estimates obtained by BTEM from near-infrared and infrared spectra. Target partial least squares has been shown to take advantage of the partial least squares regression framework in a way that uses pure component spectra as a property vector to estimate relative amounts of minor components and avoid contributions from background \cite{TPLS}. 

The quantitative efficacy of this combination can be compared with that of MCR-ALS, a standard technique for modeling pure component spectra and concentration profiles from infrared spectra \cite{irex}, hyperspectral images \cite{nirex}, and a variety of other applications \cite{MCRRev}. MCR-ALS was selected as a base of comparison for this study because it has been presented for the quantitation of trace \cite{traceconc, tracenotquant} components in mixtures. 

MCR-ALS and other curve resolution methods have not been demonstrated to work well in resolving components with relatively small signatures or components in mixtures that have poorly represented background contributions \cite{itffa, tracenotquant}. Components that have low concentrations can have relatively large signals due to their instrumental responses. Previous MCR-ALS studies related to analytes at low concentrations did not address the case where the low concentration components also had low signals for quantitation \cite{traceconc, tracenotquant}. MCR-ALS estimates cannot be obtained because collecting data with sufficient independent variation of each component in a given mixture is not possible due to a lack of experimental control \cite{selena1}. Components that suffer from these conditions are said to have poor representation because they cannot be adequately sampled for resolution. A primary goal of this study is to investigate the efficacy of BTEM and T-PLS for resolving spectra and obtaining semiquantitative estimates under conditions where MCR-ALS estimates cannot be readily obtained.

This study first shows that T-PLS can be used to construct models that are more accurate than individual CLS estimates obtained from BTEM recovered signals, and that these models can be competitive with those from MCR-ALS on a dataset where the response of the components were each relatively large and where there were minimal background effects. Then, the utility of BTEM and T-PLS for obtaining pure component estimates is investigated for situations where the spectral signatures of the pure components are low in magnitude and where sufficient representations of every analyte are difficult to attain. The efficacy of the hybrid method used under these more challenging conditions is demonstrated with two experimental datasets in which BTEM and T-PLS succeeded at modeling the pure components known to be present in those mixtures, but MCR-ALS failed. 




\section{Theory}
\subsection{Multivariate Curve Resolution Alternating Least Squares}
The goal of multivariate curve resolution (MCR) is to obtain estimates for the pure component spectra ($S$) and the respective relative concentration profiles ($C$) for the components of an unknown mixture. Because this study focuses on spectroscopic responses obtained from chemical mixtures, the relationship between the matrix of measured absorbances ($A$) and the pure components can be considered as an application of Beer-Lambert-Bouguer law where $A = CS^T$. Given some absorbance matrix $A$, MCR can be used to model candidate solutions for $C$ and $S$. Due to inherent ambiguities in posing this inverse problem, however, there is often not an exact solution for $C$ or $S$ \cite{ambiguity}. Instead, practitioners seek chemically plausible solutions.

Multivariate curve resolution alternating least squares has been shown to provide, in many instances, satisfactory empirical estimates for both the spectra of pure components, and their respective concentration profiles \cite{MCRRev, MCRorig}. Many variations of MCR-ALS have been presented in the literature. One commonality shared across all of them is that all estimated pure components present in a mixture are modeled simultaneously.

Alternating least squares solves the inverse Beer-Lambert-Bouguer problem through iteritive least squares projections of both undetermined matrices onto one another,
\begin{equation} C = A (S^T)^{-1} \end{equation}
\begin{equation} S = C^{-1} A \end{equation}
The matrix inverses of both $S$ and $C$ are often singular due to rank deficiencies in the spectral data. As a way to avoid rank-related difficulties, the Moore-Penrose pseudoinverse is commonly employed \cite{MCRorig}.

Because MCR-ALS decomposes the spectral responses in the $A$ matrix based on all of the plausible information in $C$ and $S$, an important factor in building an effective MCR-ALS model is the selection of an appropriate number of pure components. Without an adequate estimate of the number of components in the mixture, the pure component spectra and concentration profiles often become unreliable \cite{reliable}. Domain knowledge can be used to provide reasonable estimates for the number of pure components. When domain knowledge is not available, the number of components for a MCR-ALS model can be selected heuristically by a Scree plot of the eigenvalues that result from the singular value decomposition of the mixture matrix $A$ \cite{scree}. 

Even if an appropriate number of pure component estimates has been selected, ambiguous solutions may still be obtained from the nonconvex optimization of $C$ and $S$\cite{mcralsambig}. Many schemes have been employed to constrain the alternating least squares solutions to chemically plausible regions. The most commonly used constraints are those based on physical laws such as nonnegative concentrations and mass balance. Constraints based on domain knowledge such as unimodality have also been applied \cite{MCRorig}. Perhaps the simplest way to reduce ambiguity in equations 1 and 2 is by selecting regions of interest in the mixture spectra, a method sometimes referred to as band targeting \cite{BTEM}. 

Starting from suitable initial estimates for $C$ and $S$ is another common way to obtain useful MCR-ALS models. Evolving factor analysis and SIMPLISMA are two curve resolution methods that have been used to provide reasonable initial estimates of the pure component spectra in matrix $S$\cite{toolbox}. Many of the other curve resolution methods are now commonly used as starting estimates for iterative algorithms like MCR-ALS rather than as stand alone techniques \cite{selena1,selena2,toolbox}.

\subsection{Band Target Entropy Minimization}
Band target entropy minimization is another spectral recovery technique that may be used to model estimated pure spectral components, but unlike MCR-ALS, it does not simultaneously estimate concentration profiles. BTEM is used to resolve pure component spectra from mixtures through the use of singular value decomposition (SVD) of the $A$ matrix. Singular value decomposition changes the basis of a matrix such that the mutually orthogonal axes that contain the most variance are contained in the loadings matrix ($V$), the square magnitude of variance explained by the loadings is contained in the diagonal entries of the axis of $S$, and the scores matrix ($U$) are populated with the projections of each observation vector in $A$ onto the loading axes.
\begin{equation} A = U S V^T \end{equation}

BTEM utilizes an optimization technique, typically simulated annealing, to find an estimated spectrum vector, ($\hat{a}$), whose normalized first (or higher order) derivative with respect to wavelength ($\lambda$) has minimal Shannon Entropy ($H$), which is defined for any probability value($p$) as follows, 
\begin{equation}  H(p) = -p \sum \log_2(p) \end{equation}

The use of Shannon entropy in the calculation of the BTEM objective function is not rigorously founded because derivatives of spectra are not stochastic vectors following the laws of probability. However, by normalizing the entries $ |\frac{d\hat{a}} {d\lambda}|$ by their maximum value, the spectral derivatives are constrained between zero and one, similar to a probability vector for a stochastic process. The core idea behind the use of the entropy argument in BTEM is that it allows for the recovery of a spectrum that is minimal in differential information through a projection of the loadings. Those spectra are attained by minimizing the objective function($O$), where $O = argmin H( |\frac{d\hat{a}} {d\lambda }| )$,
and where $\hat{a}$ is defined as
\begin{equation}  \hat{a} = t S_{trunc} V_{trunc}^T \end{equation}
\noindent In Equation 5, $t$ is the vector obtained from the optimization, and $S_{trunc}$ and $V_{trunc}^T$ are the singular values and their respective loadings that contain a component of interest, respectively. The singular values included in the BTEM model are usually decided upon by band-targeting spectral regions from loading plots obtained by SVD that appear to contain a pure component \cite{BTEM}. In our experiments, however, using the entire spectrum rather than band targeting select regions was often equally effective ($S_{trunc} = S$ and $V_{trunc} = V$). Typically, the estimated pure spectral response $\hat{a}$ is constrained to be non-negative based on heuristic rules. The computational aspects and suggestions for implementation of the nonnegativity constraints used in BTEM can be found in greater detail in \cite{BTEM}.

In essense, BTEM is used to find a linear combination of loadings obtained from SVD that gives a smooth, simple representation. Such simple representations have been previously shown to be effective estimates of the spectral responses of pure components in mixtures \cite{BTEM,BTEMex}. The most notable limitation to BTEM curve resolution is that the algorithm requires that each pure analyte must vary independently enough relative to the other components in the mixture for it to be reasonably represented by linear combinations of the loadings matrix obtained from the singular value decomposition\cite{BTEM}. %

Garland et al. quantified the pure component spectral estimates obtained from BTEM by using the classical least squares regression framework\cite{BTEM}. The equation for the relative quantification of BTEM recovered spectra is,
\begin{equation}  \hat{C} = A \hat{a}^T(\hat{a} \hat{a}^T)^{-1} \end{equation}
Equation 6 is equivalent to replacing the regression weights in a typical classical least squares model with a pure component spectral estimate. Typically, classical least squares modeling is used to relate multivariate instrumental responses to linear changes in the amount of the pure component spectra, but, for CLS to give rank sufficient solutions, the spectra of all pure components present in the mixture must be included in the model. Equation 6 has been used in earlier work \cite{BTEM}, on a single recovered spectrum, or a vector, which implies that the equations intended use is for pure spectral regions. However, if the response of a component in a mixture does not have a pure region, as is common in near infrared spectroscopy, there is no direct extension of Equation 6 to use on multiple recovered spectra. Users of BTEM are limited to the quantification of each component one-at-a-time knowing that rank arguments are not satisfied. This requirement imposes significant limitations on BTEM quantification.

\subsection{Target Partial Least Squares Regression}
Calibration of one component in a mixture in the absence of property information is a challenge. An alternative calibration method, target partial least squares regression (T-PLS), first introduced by Feudale, et al. allows for the semiquantitation of a single pure analyte spectrum in a mixture of components \cite{TPLS}. Target partial least squares regression employs the same algorithm as conventional partial least squares regression, with a few changes. Partial least squares regression describes the relation between vector of property values, $y$ and a matrix $X$ through a reduced rank linear model that maximizes the covariance of $X$ and $y$ \cite{PLSreg}. The major difference between the conventional PLS and T-PLS algorithms is that the T-PLS algorithm is used to project $y$ into the observation space of $X$, rather than seeking latent features as a result of the projection of $y$ onto the variable space of $X$. The $y$ vector for T-PLS is a pure spectrum of a target component that is hypothesized to be present in a mixture rather than a latent property. In this work, we refer to these projections as latent projections rather than latent variables, although the mathematics are similar. The only adjustments required to allow for the nonlinear iterative partial least squares (NIPALS) algorithm to be used for the creation of T-PLS models is that the calculation of the weight matrix (W) and all subsequent calculations must be transposed accordingly. The NIPALS algorithm for T-PLS, which follows the same nomenclature as that of traditional PLS \cite{PLS}, is provided in Algorithm 1.

\newcommand{\commentsymbol}{//}
\newcommand{\LineComment}[2][\algorithmicindent]{\Statex \hspace{#1}\commentsymbol{} #2}
\begin{algorithm}
	\caption{Target Partial Least Squares}\label{Target Partial Least Squares Pseudocode}
	\begin{algorithmic}[1]
		\Procedure{  }{}
		\State u = y
		\State for(L in 1 to LatentProjections)
		\State \ \ \ \ while($||t_{new} - t_{old}||_2 < tolerance$)
		\State \ \ \ \ \ \ \ \ $w = || uX^T ||_2$
		\State \ \ \ \ \ \ \ \ $t_{old} = t_{new}$
		\State \ \ \ \ \ \ \ \ $t = X^T w$
		\State \ \ \ \ \ \ \ \ $q = 1$
		\State \ \ \ \ \ \ \ \ $p = \frac{t_{new}^TX}{t_{new}^Tt_{new}}$
		\State \ \ \ \ end while
		\State \ \ \ \ $b = \frac{ut}{t^Tt}$
		\State \ \ \ \ $X_L = X_{L-1} - pt^T$
		\State \ \ \ \ $Y_L = Y_{L-1} - Bt^T$
		\State \ \ \ \ $SSQ_X = \sum_j \sum_i (\frac{X_L}{X_{L-1}})^2 $
		\State \ \ \ \ $SSQ_y = \sum_j \sum_i (\frac{y_L}{y_{L-1}})^2 $
		\State \ \ \ \ Append w, t, q, and p vectors to matrices W, T, Q, and P
		\State \ \ \ \ Append b to a vector B
		\State end for
	 	\LineComment[1\dimexpr\algorithmicindent]{Where $\delta_{i,i}$ is the Kronecker Delta}
		\State $m_{k,} = \sum_i^k (W_{i,} (P_{,i}^TW_{i,})^{-1} (b_{i} \delta_{i,i}))$
		\EndProcedure
	\end{algorithmic}
\end{algorithm}

Although T-PLS and PLS models are created from similar algorithms, the interpretations of these models and their validation processes are different \cite{TPLS}. This is because a pure spectrum is used in place of the property vector and the regression coefficient ($m$) may be used to provide relative abundances of the target. The optimal number of latent projections for T-PLS models cannot be obtained from cross-validated results, as is common for most PLS models, because T-PLS is designed for use when property values, such as concentrations, are not known. A plot of the percent variance explained in $X$ or $y$ calculated from the residual sum of squares ($SSQ$) vs number of latent projections may be used decide on the number of latent projections used in the model. This number can be selected by the point of diminishing return \cite{TPLS}, similar to K-means clustering or by setting a threshold on the number of projections required to account for a set percentage (e.g. 95\%) of the explained variance. This approach is most useful if the noise of the measurements is either known or estimable. 
\section{Data}
\subsection{Triliquid Data}
The triliquid dataset is a designed, near-infrared data set collected at wavelengths of 1,100-2,500 nm using a FOSS 6500 instrument in transmission mode. The dataset is composed of spectra of acetic acid, methanol, and water at volume fractions of 0, 25, 50, 75, and 100\%. Every sample in the design was prepared and measured in duplicate \cite{triliquid}. 

For the curve resolution comparison experiments, the samples that contained 100\% concentration of a single liquid were not included in the training data. These pure samples were used only for comparisons with the pure component spectral estimates and calibration methods.

\subsection{Milk Adulteration Data}
The milk adulteration dataset was collected with a diffuse reflectance microPHAZIR\texttrademark \: near infrared spectrometer over the wavelength range of 1,595.7 - 2,396.3 nm. The purpose of this dataset is to determine whether a sample is milk powder (48 samples) or milk powder adulterated with melamine (29 samples).

\subsection{Vapor Release Data}
The vapor release dataset consists of infrared hyperspectral measurements of dimethyl methylphosphonate (DMMP) vapors released from a gas stack at 170$^{\circ}$C. The release was measured against a fixed background at a distance of 1.5 km using a hyperspectral infrared spectrometer created by Physical Sciences Inc. The infrared spectrometer measured spectral responses over a range of 1,270 - 920 cm$^{-1}$ at a spectral resolution of 10cm$^{-1}$. The hyperspectral images collected from the imaging spectrometer formed a data cube of 64 by 64 spatial pixels with 36 spectral measurements obtained at each spatial pixel. A black body radiation correction was applied because the excitation source used here was ambient sunlight \cite{TPLS}.

\section{Results and Discussion}

\subsection{Comparison with Known Methods}
The triliquid data is a designed set of three polar liquids with volume fractions that varied in 25 v/v \% increments. Because this dataset had known chemical composition and minimal contributions from background, it was used for assessing the BTEM estimation of pure spectral components and the accuracy of quantification for each known component in the mixture.   

The pure component spectra for water, acetic acid and methanol were estimated from the mixture spectra using band targeting and entropy minimization. The pure component spectral estimates obtained from BTEM were subjected to a second-order Savitsky-Golay smoothing filter to remove noise artifacts, as can be seen in Figure 1. The parameters of the Savitsky-Golay filter were manually tuned until the spectra resembled smooth line shapes. Procrustes distance analysis was employed to compare the shape similarities of the resolved pure component spectra and the experimentally-obtained, pure component spectra after rescaling the smoothed spectral estimates by their maximum value. The Procrustes distances from the Savitsky-Golay smoothed BTEM estimated pure component spectra to the experimentally obtained pure spectra of methanol, acetic acid, and water were 0.758, 1.32, and 1.66 rescaled absorbance units, respectively. 

\begin{figure}[h]
	\centering 
	\includegraphics[width=0.95\linewidth]{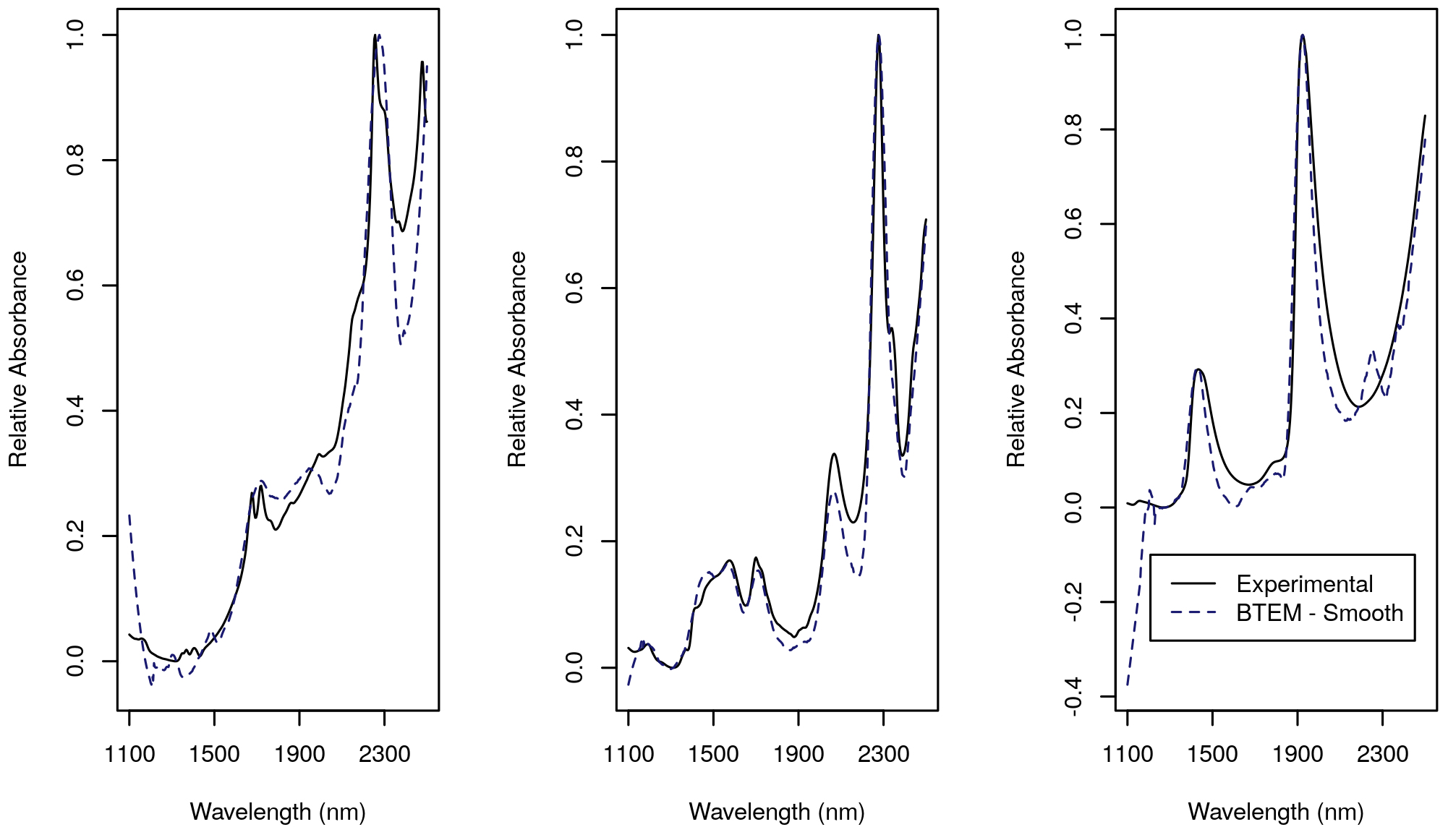}
	\caption{Rescaled experimental and smoothed BTEM estimates of acetic acid (left), methanol (center), and water (right) spectra. }
\end{figure}

By the Procrustes distance metric, the smooth, BTEM-recovered methanol spectrum had a shape that was most similar to its experimental spectrum. The reason why the BTEM methanol estimate and experimentally collected methanol spectrum were roughly two-fold more similar than acetic acid and water were to their respective experimental spectra may have been methodological, but it may have also been a result of the chemistry of the mixtures studied. In both the resolved components for acetic acid and water, there were relatively large distances from the experimental spectra at small wavelengths that may have been artifacts of the BTEM curve resolution method. It is known that shifts in vibrational energy needed to excite the transitions in acetic acid can be bought on by the interaction between polar components of mixtures, dimerization, and because of proton dissociation equilibria \cite{genchem}. Because of these chemical interactions, the BTEM estimates of acetic acid and those of water likely can be expected to not match the experimentally-obtained pure acetic acid spectra 

Some evidence that suggested that different or shifted species existed for acetic acid was the presence of a band at approximately 2300 nm in the BTEM estimated pure component water spectrum. This claim was assessed by performing principal components analysis on the data in the 2250 - 2300 nm range and a second PCA excluding that range using only 3 principal components, without any sample replicates. Ideal simplex experiments form linear simplices in the scores space of a PCA \cite{triliquid}. Figure 2 demonstrates that the PCA formed by using only the spectral information from 2250 - 2300 nm was more nonlinear than that obtained from the remaining wavelengths. The scores of the first three principal components of the band of interest had a coefficient of variation for the nearest neighbor distances of 235\%. The coefficient of variation obtained from the wavelengths without the 2250-2300 nm band was only 163\%. This indicated that the spacing between the points of the simplices in the scores space were less dispersed, and therefore more linear. Because the experimental design should be linear at rank 3 it is likely that the band is representative of an interaction effect but it could also be explained as an artifact of a peak in acetic acid or methanol such as a CH$_3$ combination band that covaried with the bands attributed to water.

\begin{figure}[!h]
	\centering 
	\includegraphics[width=0.95\linewidth]{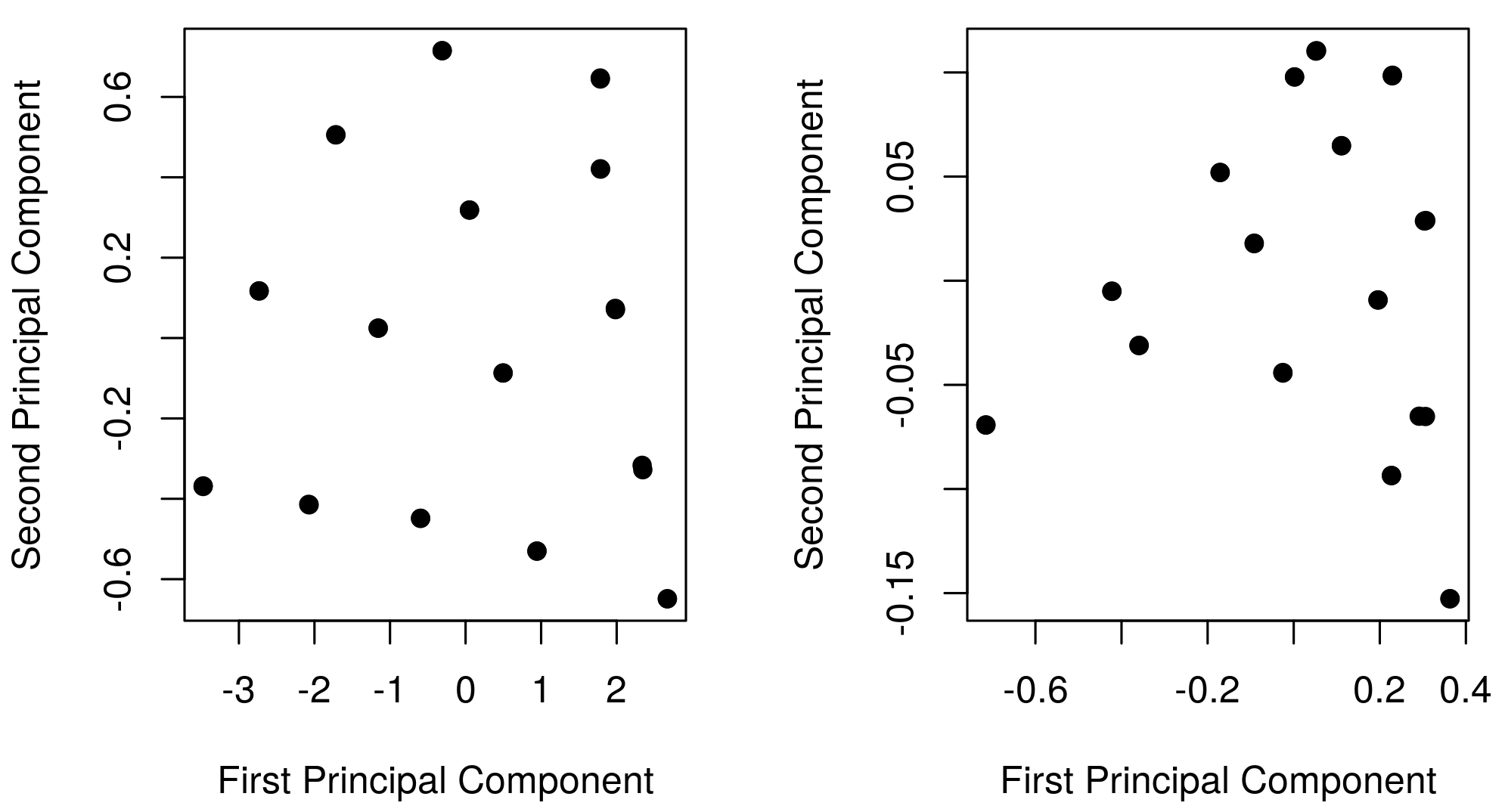}
	\caption{Scores plots of the triliquid data of the first two principal components of the triliquid data without the absorbance measurements obtained at 2250 - 2300 nm (left), and of only the absorbance measurements obtained in the region of 2250 - 2300 nm (right). }
\end{figure}

Due to potential interactions between the component species in the triliquid mixture and possible methodological errors present from BTEM curve resolution, a detailed comparison between experimentally obtained and curve resolved pure component spectra could not be made. Our intended usage of the pure component spectral estimates obtained from BTEM was semiquantitative analysis. This dataset featured known concentration values so that quantitative efficacy could be explored in a way similar to that done with typical calibration models. The predictive error was calculated for each component by range scaling \cite{rangescaling} the predictions between 0-100\% and comparing them with the known v/v \% amounts. The root mean squared errors of estimation (RMSE) obtained from CLS and T-PLS regression models built from experimental, BTEM resolved, and second-order Savitsky-Golay smoothed BTEM resolved pure component spectra were compared with results attained from MCR-ALS quantification with three pure components, as shown in Table 1. 

\begin{table}[H]
	\centering
	\begin{tabular}{ccccc}
		\hline
		Curve Resolution & Calibration & Acetic Acid & Methanol & Water  \\ 
		Method & Method & (\% RMSE) &  (\% RMSE) & (\% RMSE)  \\ 
		\hline
		None  & CLS & 36.43 & 35.31 & 30.28 \\ 
		(Experimental)& T-PLS & 12.72 & 4.89 & 5.66 \\ 
		\hline
		BTEM & CLS & 44.89 & 44.27 & 49.55 \\ 
		(raw) & T-PLS &  18.24 & \textit{6.19} & \textit{9.78} \\ 
		\hline
		BTEM & CLS & 47.27 & 45.96 & 48.53 \\ 
		(smoothed) & T-PLS & \textit{11.73} & \textbf{4.69} & \textbf{9.75} \\ 
		\hline
		MCR & ALS & \textbf{7.19} & 16.51 & 17.25 \\ 
		\hline
	\end{tabular}
	\caption{ Percent root mean squared errors of range scaled relative concentration estimates obtained from classical least squares, target partial least squares, and MCR-ALS for the known pure components in the triliquid dataset. Entries in bold text represent the lowest errors of the curve resolved calibrations, and those in italics are the entries which had the second lowest curve resolved errors. }
\end{table}

As expected, the CLS regression models of single pure components all produced the largest predictive errors. The failure of CLS to adequately model single components is in agreement with the theoretical understanding that CLS requires full rank regression weights and that individually regressing each component of a three-component mixture gives rise to a rank deficient solution. The T-PLS models that were built from the smoothed BTEM pure component spectra tended to yield the lowest errors of estimation. 

Although the calibration of curve resolved spectra was the focus of this study, it was interesting that the errors that resulted from T-PLS models that were calibrated using the smoothed, BTEM-resolved spectra for acetic acid and methanol were lower in magnitude than those obtained from the experimentally collected spectra. The finding that these curve resolved components had lower errors of estimation than those from experimentally obtained pure spectra supports our claim of chemical interactions in the triliquid mixture.

In this experiment, the target partial least squares models built from the three smoothed BTEM pure component spectra had lower or nearly commensurate errors as those from the three-component MCR-ALS model. These MCR-ALS models were built using spectral non-negativity and closure constraints. Many more varieties of MCR-ALS models were investigated; details of these are omitted from this report, but in our experiments, no three component MCR-ALS model had lower predictive errors for methanol and water than those from BTEM and T-PLS.

MCR-ALS models with 1, 2, 4, and 5 pure components were also built using spectral non-negativity and closure constraints so that an assessment of the variability of the MCR-ALS models could be undertaken. The errors of estimation for these experiments are displayed in Table 2. The one component MCR-ALS model only resolved the water spectrum, likely because the NIR water bands are of greatest intensity in these spectra. It is common for curve resolution methods to resolve the component which has the greatest intensity first \cite{itffa}. On the triliquid dataset, it was possible to manually recover individual spectra of pure components with BTEM regardless of a component's relative intensity, provided the experimental spectra were band targeted appropriately. The two component MCR-ALS model featured a pure component spectrum with bands that were present in both the experimentally obtained acetic acid and methanol spectra; thus, the errors of prediction for both analytes were assessed for both components using the one spectrum. The two component MCR-ALS model was unable to exclusively recover the spectra of either methanol or acetic acid, although a 3 component MCR-ALS model was able to do so. 

For this dataset, the Scree plot indicated that there were three or four pure components in the mixture. In the case of three components, BTEM and T-PLS calibration resulted in lower predictive errors for water and methanol, but not for acetic acid, as stated previously. The four component MCR-ALS model, however, had lower errors of estimation for water and methanol, but not for acetic acid. The errors of estimation for methanol from the four component MCR-ALS model was only 10\% different than the T-PLS calibrated BTEM estimate. The variability between estimation errors obtained with MCR-ALS by choosing 3 or 4 components in the mixture resulted in a difference of 10.06\% RMSE for acetic acid. The difference in quantitative efficacy between both models with the selection of a single parameter is noteworthy. Overall, the MCR-ALS models with three or four components resulted in errors that were similar to those obtained from the BTEM and T-PLS modeling.  

\begin{table}[H]
	\centering
	\begin{tabular}{cccc}
		\hline
		Pure Components & Acetic Acid & Methanol & Water  \\ 
		(\#) & (\% RMSE) &  (\% RMSE) & (\% RMSE)  \\ 
		\hline
		1 & N/A & N/A & 20.79 \\
		2 &  43.66 & 44.97 & 8.50 \\ 
 		4 &  15.82 & 4.20 & 7.68 \\ 
 		5 &  17.24 & 6.41 & 9.55 \\ 
		\hline
	\end{tabular}
	\caption{ Percent root mean squared errors of range scaled relative concentration estimates obtained from MCR-ALS with 1, 2, 4, and 5 estimated components in the triliquid dataset. Entries denoted by N/A indicate that concentration estimates could not be obtained using MCR-ALS.}
\end{table}

The similarities between the errors obtained from the best MCR-ALS models (3 or 4 components) and those obtained from T-PLS calibrated BTEM estimates were interesting because these two approaches define very different curve resolution experiments and have different methods of operation. Modeling with BTEM and T-PLS involves resolving a single pure component spectrum followed by the quantification of only that component. This analysis is performed independent of other components, while MCR-ALS simultaneously decomposes all components.  
 
For experiments where a training set with known pure components and concentrations are available, such as the triliquid data, finding the suitable number of components in an MCR-ALS model is usually a simple task. However, there are many experiments where the recovery and quantification of only a specific pure component, if it exists, in a mixture is the goal. Two studies where BTEM and T-PLS modeling can be used in conjunction to estimate relative amounts of one unknown chemical component in a mixture are presented below.

\subsection{Analysis of Contaminant in Milk Powder}
The first study concerns a quality control process, where the goal of the analysis is to assess the purity of a raw material and to rapidly identify contamination present in a sample. The approach is simple; it relies on variation induced by serial dilution. The idea is that newly acquired material of unknown quality can be diluted with pure stock material, and analyzed after each dilution. If any components other than the pure stock material are present, band target entropy minimization can be used to estimate their pure component spectra due to the changes in their signal intensity across the diluted samples, because the primary sources of variation should be limited to the concentration of the interferent and any artifacts of the measurement itself. To ensure that the contaminant spectrum obtained by BTEM is actually present in the sample matrix and not an artifact of the recovery method, T-PLS may then be used to construct a calibration model for the isolated, pure component spectrum and to reduce background effects present in the dilution measurements \cite{TPLS}.

The serial dilution experiment was assessed on a milk powder dataset. Five spectra that had melamine mass fractions of 0.05, 0.1, 0.5, 1, and 2\% were sampled. Band target entropy minimization of the entire spectral range using the first two singular values of the normalized spectral data resulted in the extraction of a pure component spectrum that was visually comparable to that of melamine \cite{melamine} (Figure 3). The entire spectral region was targeted in this analysis because the nature of the contaminant and its spectral signature were presumed to be unknown, and there were no visible adulterant bands in the NIR spectra. 

A T-PLS semi-quantitative calibration using two latent projections created from the second-order Savitsky Golay smoothed BTEM pure component spectral estimate yielded a 6.59\% relative RMSE (0.10 w/w \% absolute RMSE) over the same five sample range. This analysis was performed twice more to assess variability, using different samples that had the same diluted concentrations. A mean relative RMSE of 11\% was calculated. The R-squared values obtained from linear fits for the actual versus predicted concentration plots generated from the T-PLS calibrations were all $>$0.90, which indicated a linear calibration. The linearity of the T-PLS calibrations were lower than what might be expected from an ordinary partial least squares regression, but still strongly supported the claim that the component recovered by BTEM was not a spectral artifact, and was in fact a component that varied proportionally with the dilution process. 

MCR-ALS modeling with SIMPLISMA-estimated initial pure component spectra was also applied to the same samples used in the BTEM T-PLS modeling in an attempt to discover the adulterant. In our one-component MCR-ALS analysis, non-negativity constraints were applied, but the only pure component spectral estimate that was obtained was visually similar to the unadulterated milk samples, and not to melamine. Even though the SVD Scree plot indicated only one major component, we investigated a second model using two pure components. This resulted in two highly similar spectra, both of which were again representative of the unadulterated milk powder. For this dataset, MCR-ALS modeling was unable to find the melamine contaminant, nor to quantify it. We attribute the failure of the MCR-ALS models to recover the pure components associated with melamine to the relatively low intensity of the contaminated response, and to the scatter present in the reflectance measurements.

\begin{figure}[!h]
	\centering 
	\includegraphics[width=0.95\linewidth]{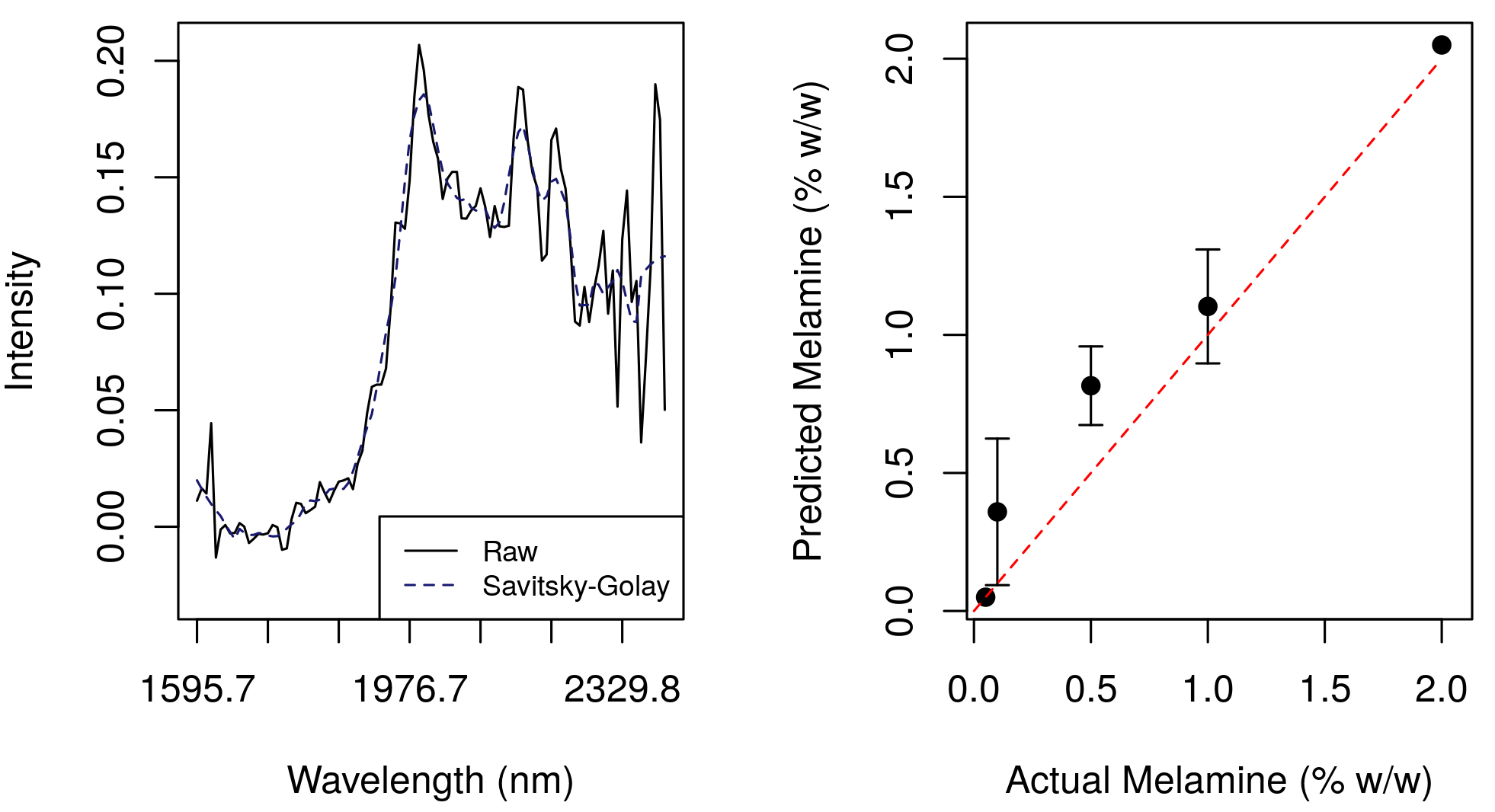}
	\caption{The raw and Savitsky-Golay smoothed BTEM estimated pure component spectra obtained from the serially diluted adulterated milk samples (left). An actual versus predicted plot of melamine concentrations obtained from the target partial least squares model using the Savitsky-Golay smoothed spectral estimate (right). The average predicted values are shown with 1 sigma error bars (N $= 3$) for their scaled predictions.}
\end{figure}

BTEM pure component spectral estimates obtained from samples that were reported to contain only pure milk powder were employed as a control. The spectra recovered from these experiments were most similar to the unadulterated milk powder spectra themselves, as can be observed in Figure 4. Calibration by T-PLS was not attempted because no other components appeared in the loading plots obtained from the singular values, and the majority of the variance across samples was attributable to noise and scatter effects. This result was repeated two more times, and each time no components were found using BTEM other than the raw milk powder spectra. In this case, MCR-ALS modeling provided the same result as BTEM; only recovered spectra which resembled the raw milk powder were obtained.

\begin{figure}[H]
	\centering 
	\includegraphics[width=0.95\linewidth]{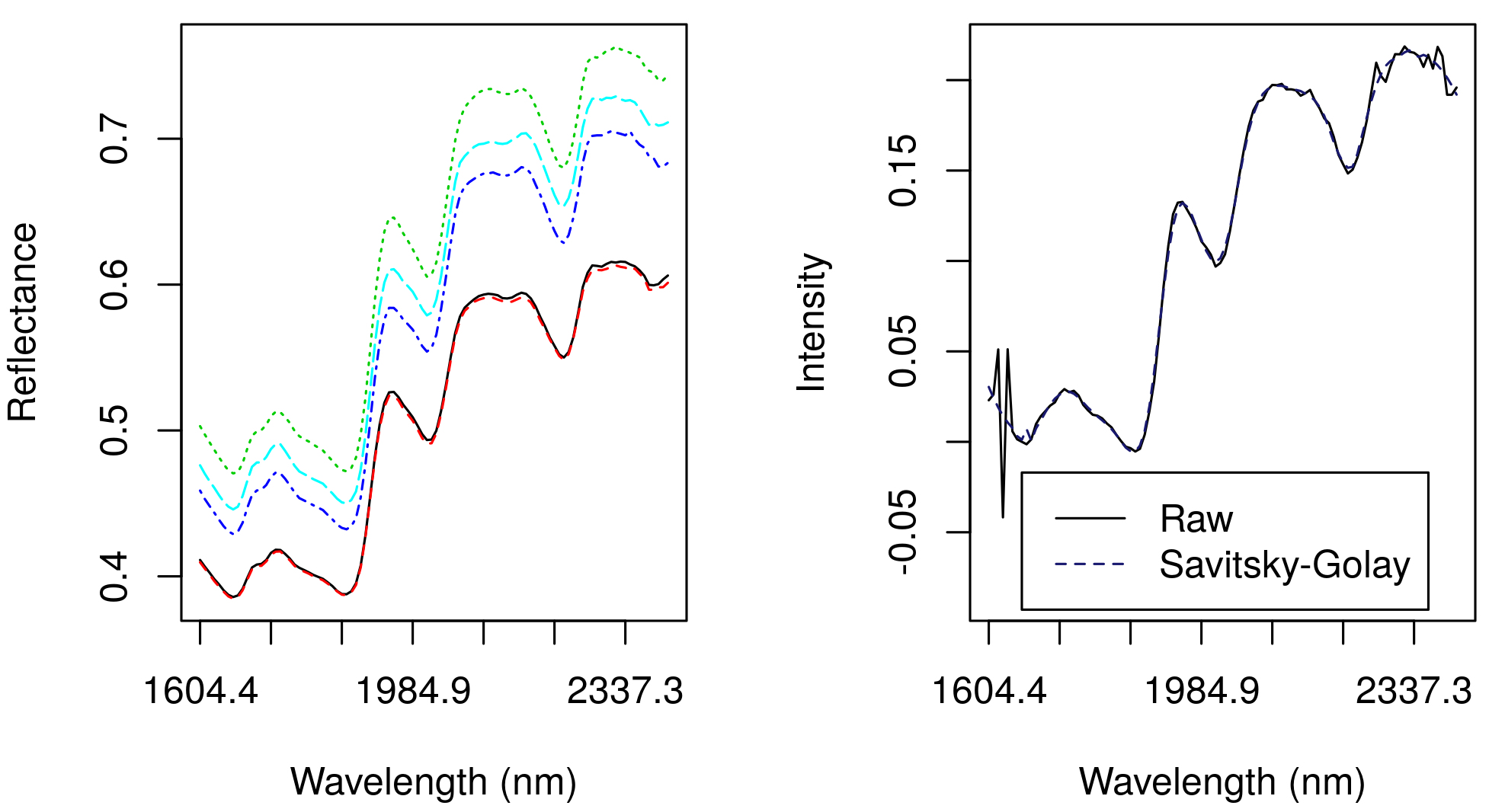}
	\caption{Overlay of five unadulterated milk powder spectra (left). The raw spectra and smoothed band target entropy minimization estimate obtained from the first two singular values of the milk powder spectra (right).}
\end{figure}

\subsection{Analysis of Hyperspectral Images}
The second study concerns the spectral recovery and calibration of chemical components obtained from concentration gradients that occur in hyperspectral images. The remote monitoring or imaging of chemicals in the environment is a challenging experimental problem. However, concentration gradients occur naturally from physical processes such as diffusion, advection, and convection. It was hypothesized that band target entropy minimization could resolve pure component spectra from natural gradients in the concentration of a target species in chemical images and that target partial least squares could be used to perform semiquantitative targeted calibrations despite a dynamic background.

The dataset we investigated for the concentration gradient study was the vapor release dataset first reported by Feudale et al \cite{TPLS}. The vapor release dataset was collected to monitor the release of dimethyl methylphosphonate (DMMP) from a gas stack via mid-infrared frequencies that are associated with phosphonate functional groups. A pure component spectral estimate was resolved from seven of the pixels near the gas stack by applying entropy minimization with three singular values using the entire spectral region. It can be seen from Figure 5 that the extracted pure component spectral response is visually similar to that of a NIST reference spectrum of DMMP \cite{NIST} which had more than an eight-fold greater spectral resolution than that used to collect the hyperspectral images. 

\begin{figure}[!]
	\centering 
	\includegraphics[width=0.75\linewidth]{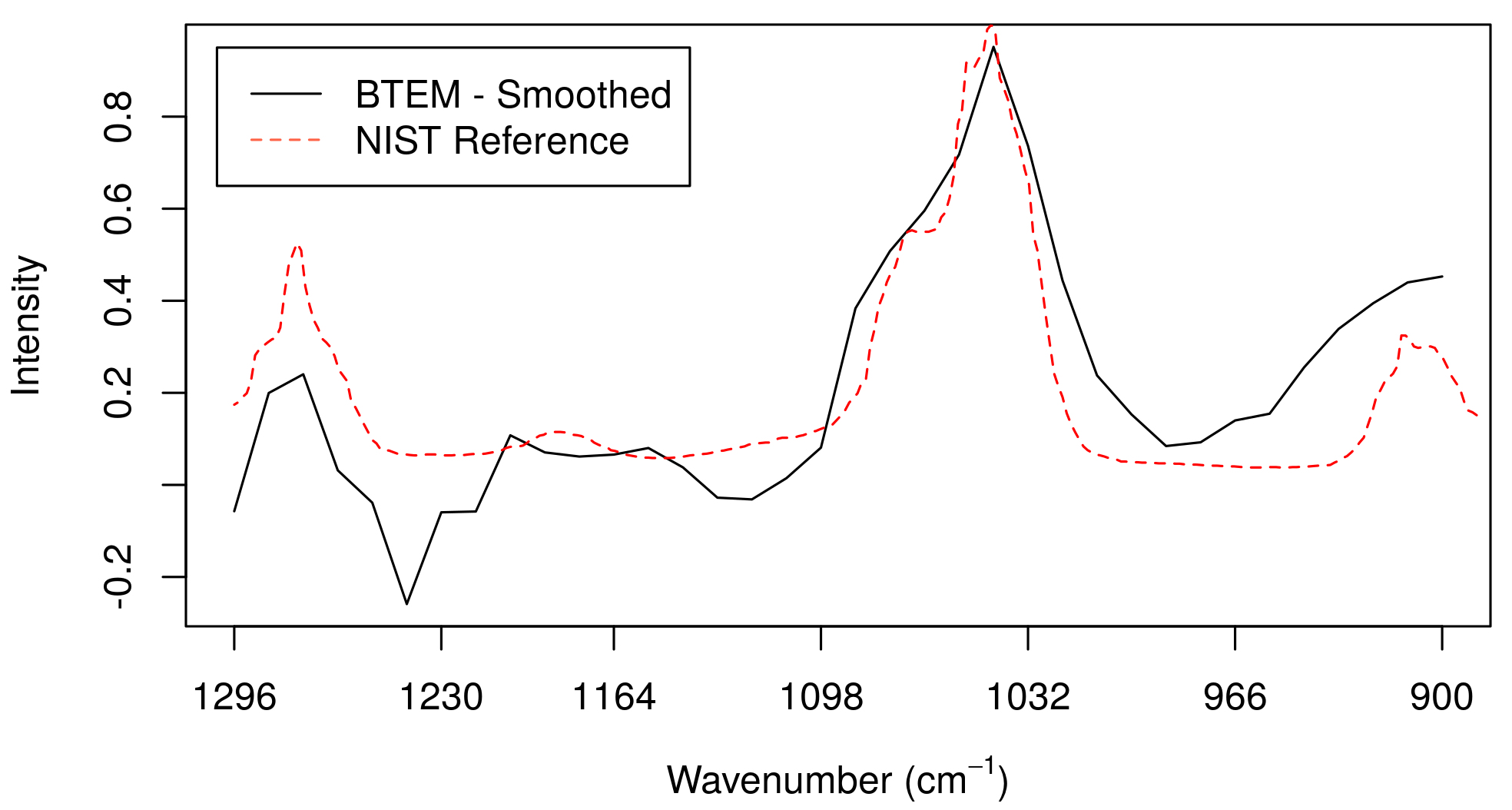}
		\begin{subfigure}{.5\textwidth}
		\includegraphics[width=.9\linewidth]{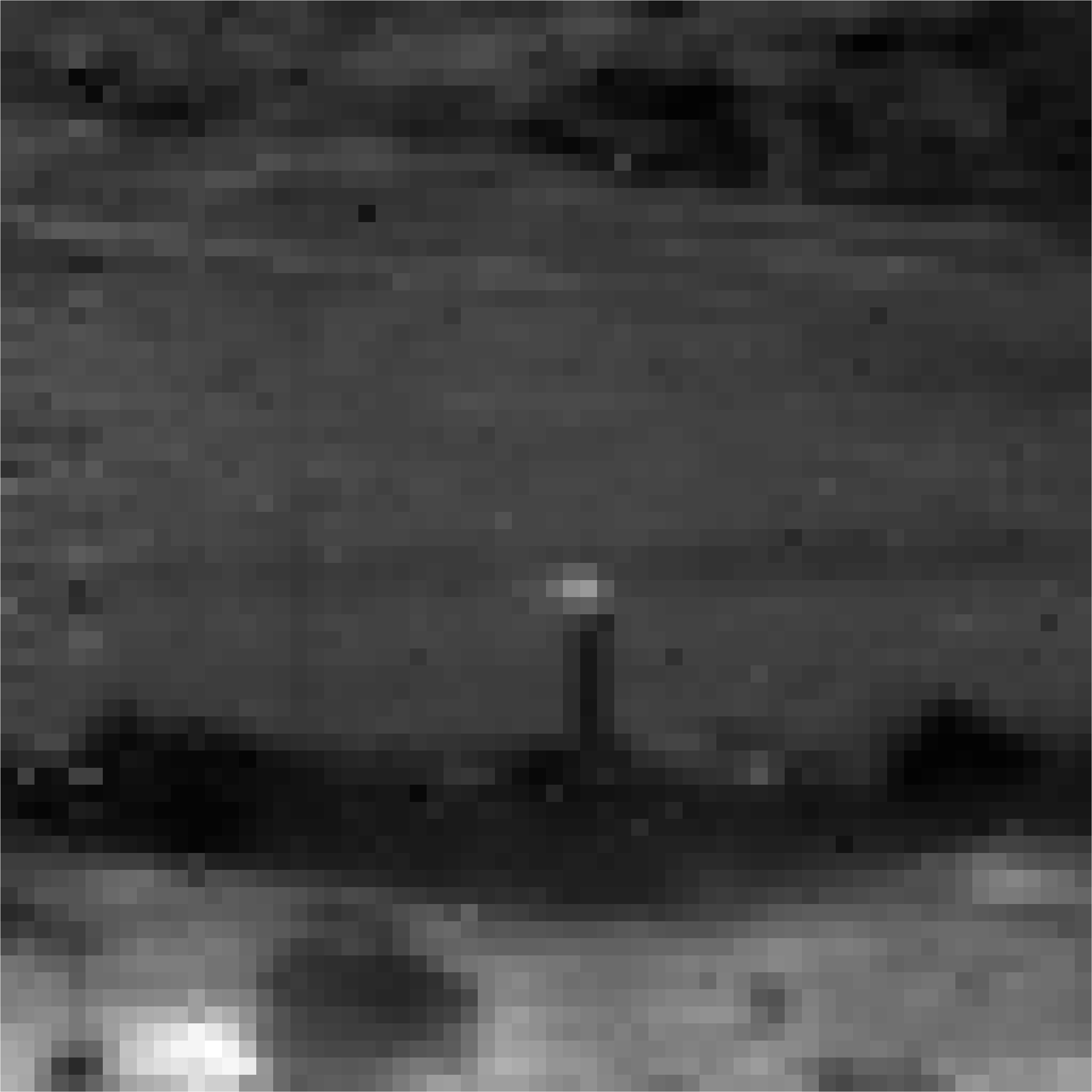}
	\end{subfigure}%
	\begin{subfigure}{.5\textwidth}
		\includegraphics[width=.9\linewidth]{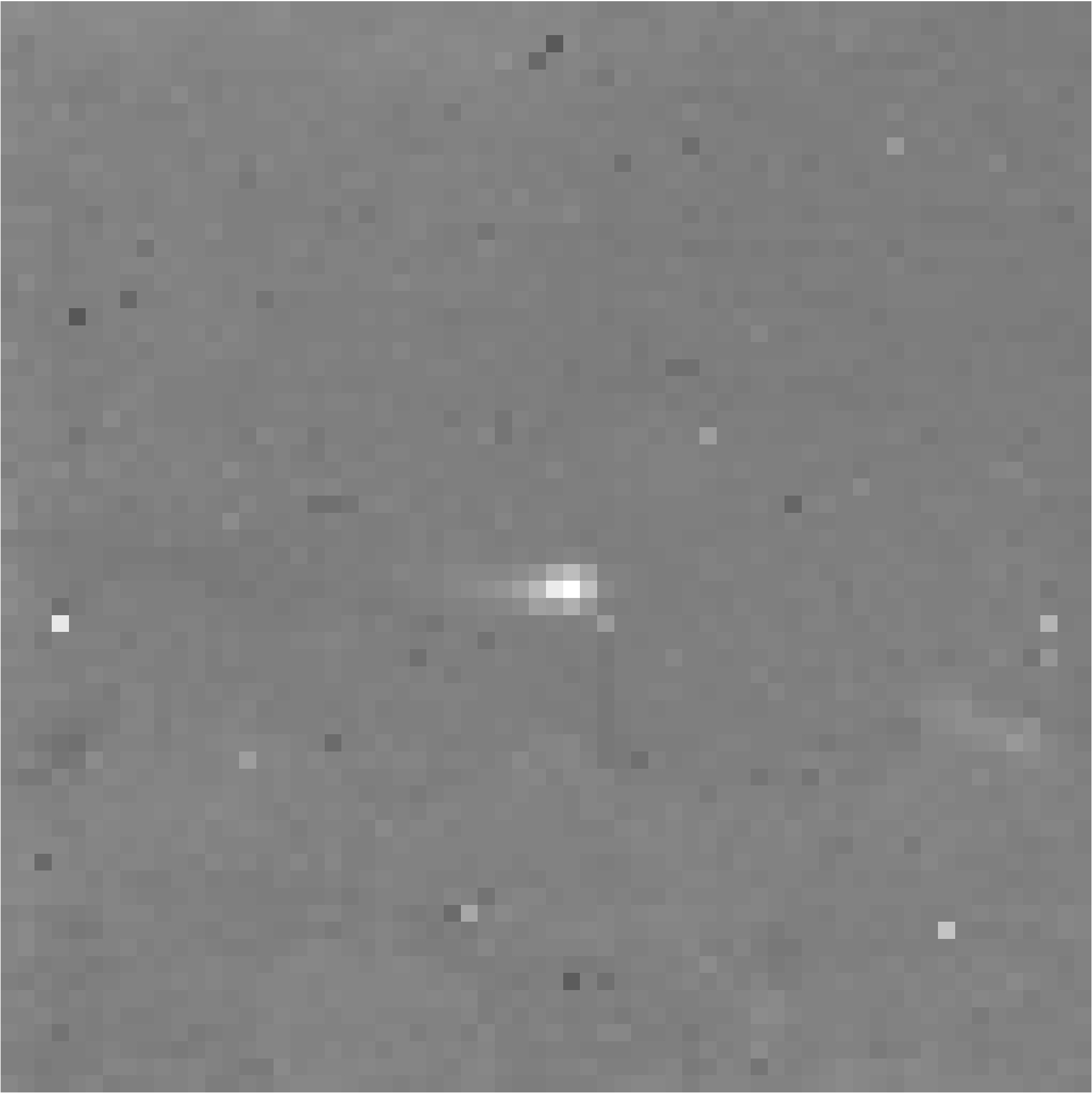}
	\end{subfigure}%
	\caption{A spectral overlay of band target entropy minimization pure component spectral estimate from selected pixels in the vapor release data and a NIST reference spectra for DMMP (top). A z-score normalized hyperspectral image of the vapor release data (bottom left). This normalization set values that were $\geq$ 2 $\sigma$ away from the mean intensity to the mean. This normalization was used only for this display and not for any of the reported analysis'. The raw heat map generated from T-PLS calibration of the BTEM pure spectral estimate of DMMP on the hyperspectral image (bottom right).}
\end{figure}

In a previous study, target partial least squares modeling was shown to be an effective method for calibrating DMMP on this dataset from a separate, experimentally-obtained spectrum of pure DMMP \cite{TPLS}. In this study, the pure component spectral estimate obtained via BTEM from the hyperspectral data was utilized as the target spectrum in T-PLS. A T-PLS calibration model was made with six latent projections (100\% variance explained in $y$, and $>$98\% in $X$) on the entire hyperspectral image using the spectral estimate obtained from BTEM. The results from the calibration are shown in Figure 6. 

The greatest estimated intensity of the recovered DMMP spectrum was found near the gas stack, similar to what was found in the previous study \cite{TPLS}. A small fraction (14/4096) of pixels appeared to have relatively large calibrated amounts of the estimated DMMP signature but were not located within a 10 pixel radius of the gas stack from which the vapor was released. It was hypothesized that those high intensity pixels were artifacts of random variation because they represented only 0.34\% of the hyperspectral image and were not part of the release experiment. To further assess the variability of the methods and those pixels, a resampling study was performed.

Jackknife mean and variance estimations\cite{jackknife} were performed to assess how variable BTEM and T-PLS estimates of DMMP concentration were for the vapor release data. Both the BTEM spectral estimates and the intensity maps obtained by T-PLS were individually range scaled between zero and one\cite{rangescaling} so that the semiquantitative T-PLS predictions could be pooled. It was found that the same pixel (X = 35, Y = 34) contained the greatest intensity for the jackknifed DMMP target estimate across all leave one out trials, which indicated uniform quantitation across the hold out trials. More evidence supporting the uniformity of the calibration was found by the fact that the estimated standard deviation heat map showed the least variation in the predictions nearest the gas stack chimney and the most variation in the background.

Interestingly, in the jackknife mean estimate, the fourteen pixels of relatively large intensity believed to be artifacts of random variations were observed to still be relatively high in intensity (Figure 7). However, those pixels also tended to have large jackknife variance estimates (relative to those pixels nearest the chimney), a result which further supported the notion that those pixels were artifacts in the data, and did not contain the target signature. This finding was congruent with an earlier T-PLS analysis on the same dataset that used an experimentally obtained target DMMP spectrum. They demonstrated that it was statistically unlikely that the pixels located away from the chimney were from the same probability distribution as the pixels near the chimney using a Kolmogorov-Smirnov test \cite{TPLS}. 

Although MCR-ALS has been shown to be a valuable tool for the analysis of hyperspectral images when the components of interest are well represented and relatively high in intensity\cite{nirex}, this dataset posed a challenge. This dataset features many contributions to variance because the spectral background depended upon the spatial location and because DMMP is a relatively minor component. MCR-ALS with SIMPLISMA initial estimates failed to recover the target DMMP spectrum using both the seven pixels selected for BTEM and the entire hyperspectral image.

\begin{figure}[H]
	\centering 
	\begin{subfigure}{.5\textwidth}
		\includegraphics[width=.9\linewidth]{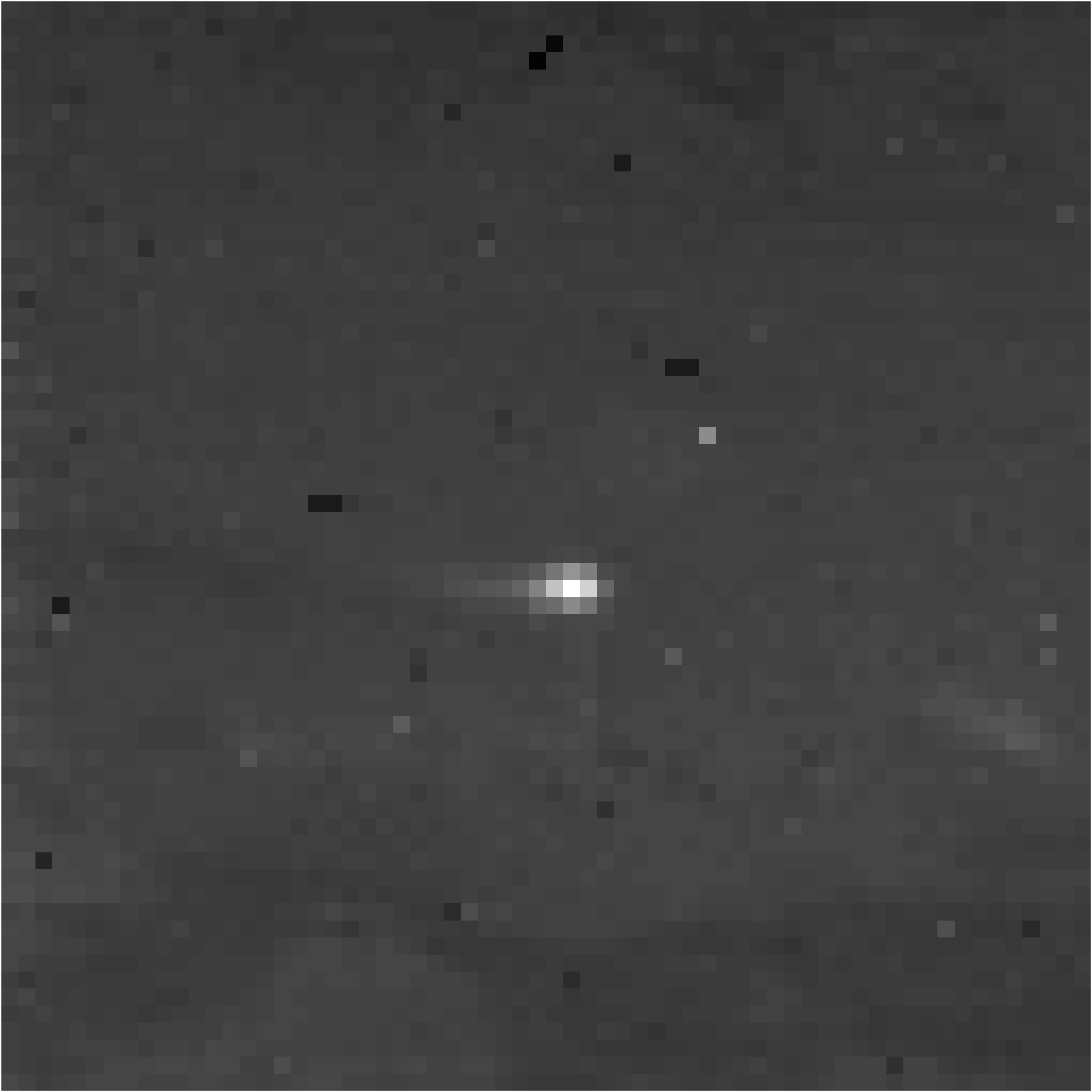}
	\end{subfigure}%
	\begin{subfigure}{.5\textwidth}
		\includegraphics[width=.9\linewidth]{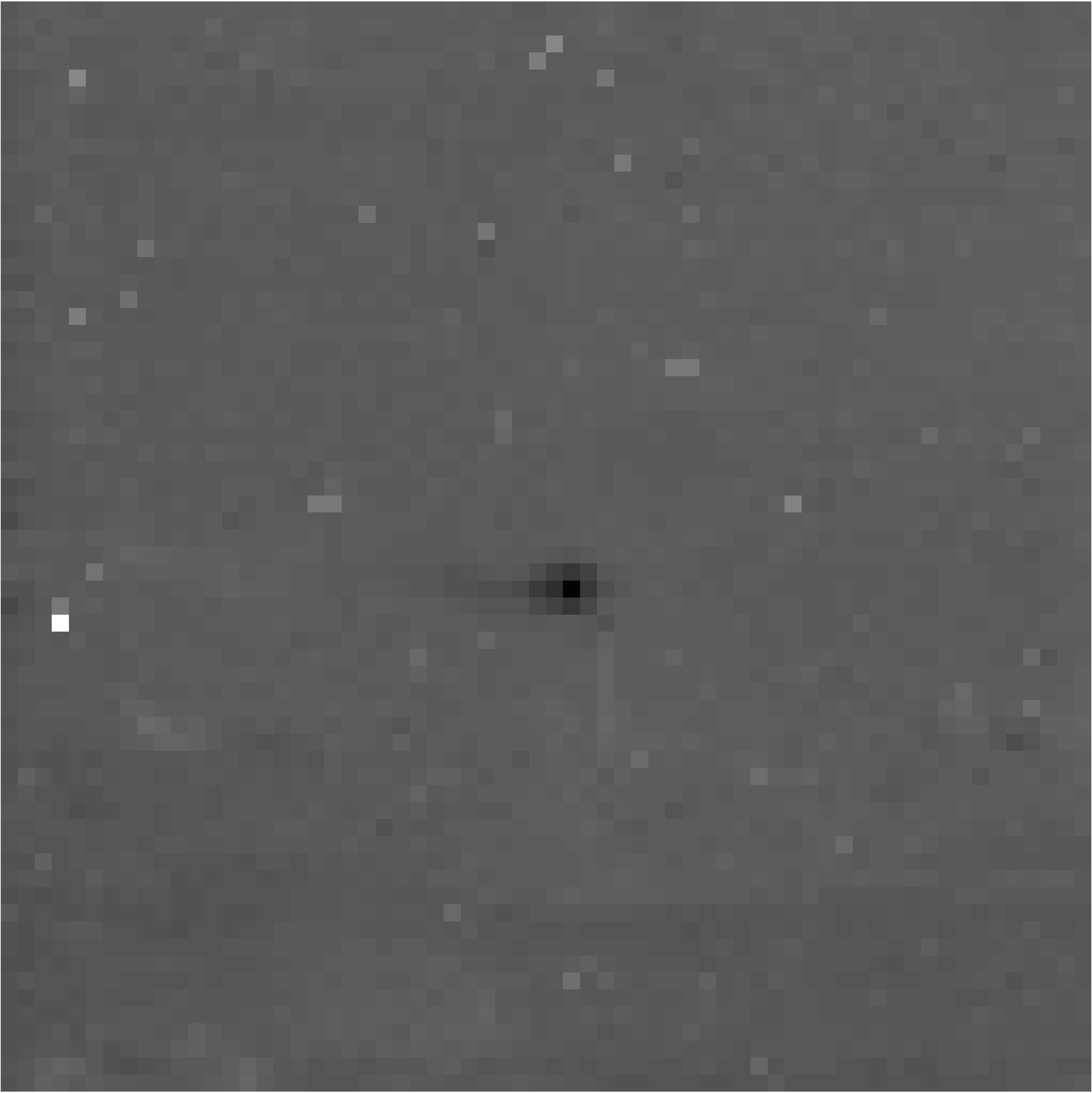}
	\end{subfigure}%
	\caption{The leave one out jackknife estimated mean (left) and standard deviation (right) intensity map estimate obtained by T-PLS and BTEM on the vapor release data (left). }
\end{figure}

\section{Conclusion}
The results from this study demonstrated that the errors of estimation obtained from target partial least squares models built by using band target entropy minimization to extract pure component spectra were lower than those from classical least squares and were similar to those from multivariate curve resolution alternating least squares for at least one dataset. It was also shown that BTEM and T-PLS can be used together to identify minor contaminant species and calibrate their presence by a simple serial dilution experiment despite scatter and noise in the measurements. Because the only major requirement for BTEM is that the signal of the species of interest varies across samples, it was also demonstrated that it is possible to extract a spectral estimate of a minor component present in the advection of vapor released from a gas stack. It was also shown that T-PLS could be used to quantify the estimated pure component spectra of the vapors present across a hyperspectral image despite the presence of spatially varying spectral fluctuations in an unknown background. 

We believe that MCR-ALS failed to estimate pure spectral components in the two mixtures due to insufficient representation of all components. The milk powder spectra appeared to be affected by scattered light and instrumental noise. Although MCR-ALS was capable of obtaining an estimate for the reference milk powder spectra, we were unable to estimate the melamine contaminant, despite the variation in its concentration presented in the samples. The vapor release dataset had unknown background effects due to the size and emissivity of the physical area which was measured and different temperatures, even after attempting to band target the vapor plume. In these cases, it was found that obtaining estimates for pure component spectra that were not linked to the success of finding the other components present in the spectra, nor their respective concentration profiles, was advantageous. Even though the mixture components were too poorly represented for MCR-ALS to obtain estimates for them, BTEM and T-PLS could be used to successfully model pure spectral estimates and to calibrate their presence. 

The combination of BTEM and T-PLS appears to be a useful one. Very few tools and strategies exist for both qualitatively and quantitatively extracting information about minor components without pure component spectra or property values. We hope that more studies which demonstrate the utility of these tools, and their variations, for hyperspectral imaging and quality control applications are performed. 

\section{Acknowledgments}
This work was supported by the United States National Science Foundation grant 1506853.

\section{Conflict of Interest}
The authors declare no conflict of interest.

\section{References}




\bibliographystyle{model1-num-names}
\bibliography{sample.bib}



\newpage

\end{document}